\documentclass{article}

\usepackage{microtype}
\usepackage{graphicx}
\usepackage{subfigure}
\usepackage{booktabs} 

\usepackage{hyperref}

\usepackage[accepted]{icml2023}

\usepackage{amsmath}
\usepackage{amssymb}
\usepackage{mathtools}
\usepackage{amsthm}
\usepackage{xcolor}
\usepackage{multirow}
\usepackage{algorithm}
\usepackage{algorithmic}
\usepackage{float}
\usepackage{tabularx}
\usepackage{booktabs}
\usepackage{amsmath}
\usepackage{makecell}

\usepackage[capitalize,noabbrev]{cleveref}

\theoremstyle{plain}

\theoremstyle{definition}

\theoremstyle{remark}

\usepackage[textsize=tiny]{todonotes}

\icmltitlerunning{A Hybrid Co-Finetuning Approach for Visual Bug Detection in Video Games}

\begin{document}

\twocolumn[
\icmltitle{A Hybrid Co-Finetuning Approach for Visual Bug Detection in Video Games}



\icmlsetsymbol{equal}{*}

\begin{icmlauthorlist}
\icmlauthor{Faliu Yi}{equal,yyy}
\icmlauthor{Sherif Abdelfattah}{equal,yyy}
\icmlauthor{Wei Huang}{yyy}
\icmlauthor{Adrian Brown}{yyy}
\end{icmlauthorlist}

\icmlaffiliation{yyy}{Xbox Studios Quality AI Lab, Xbox Game Studios, Microsoft, USA}

\icmlcorrespondingauthor{Faliu Yi}{faliuyi@microsoft.com}
\icmlcorrespondingauthor{Sherif Abdelfattah}{sherifgad@microsoft.com}

\icmlkeywords{Machine Learning, ICML}

\vskip 0.3in
]



\printAffiliationsAndNotice{\icmlEqualContribution} 

\begin{abstract}
Manual identification of visual bugs in video games is a resource-intensive and costly process, often demanding specialized domain knowledge. While supervised visual bug detection models offer a promising solution, their reliance on extensive labeled datasets presents a significant challenge due to the infrequent occurrence of such bugs. To overcome this limitation, we propose a hybrid Co-FineTuning (CFT) method that effectively integrates both labeled and unlabeled data. Our approach leverages labeled samples from the target game and diverse co-domain games, additionally incorporating unlabeled data to enhance feature representation learning. This strategy maximizes the utility of all available data, substantially reducing the dependency on labeled examples from the specific target game. The developed framework demonstrates enhanced scalability and adaptability, facilitating efficient visual bug detection across various game titles. Our experimental results show the robustness of the proposed method for game visual bug detection, exhibiting superior performance compared to conventional baselines across multiple gaming environments. Furthermore, CFT maintains competitive performance even when trained with only $50\%$ of the labeled data from the target game.
\end{abstract}

\section{INTRODUCTION}
\label{sec:intro}
Over the past decade, the video game industry has experienced substantial growth, driven by advancements in 3D game engines, increased computing power, expanded Internet bandwidth, and sustained consumer demand~\cite{game_ind_forbes}. This growth has coincided with a corresponding rise in the complexity and scale of game development, necessitating the deployment of large quality assurance teams to ensure product quality. However, the escalating demand and complexity of games present a notable challenge to this approach, as expanding the workforce entails increased costs and management overhead. One of the important game quality control activities is testing for visual bugs~\cite{taesiri2024glitchbench}, which might get introduced due to issues in the code, hardware infrastructure, or networking (e.g., player load). In general, game visual bugs could be categorized into two categories: 1) bugs detectable using a single frame (e.g., texture issues, object-to-object clipping, or object floating), and 2) bugs that need multi-frame (i.e., temporal context) to be detected (e.g., glitches and lighting issues). The latter category is more challenging for the need to consider the temporal context across multiple frames, which makes it more demanding and time-consuming for quality teams performing manual testing. 

\begin{figure*}[htbp]
    \centering
    \includegraphics[scale=0.6]{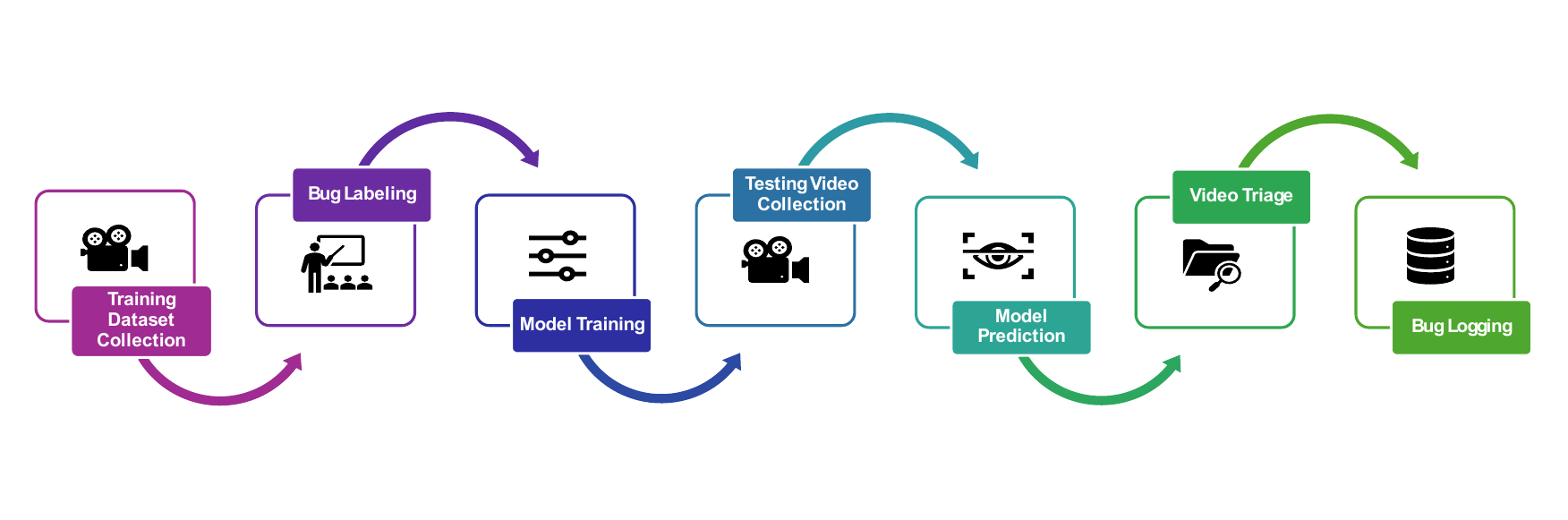}
    \vspace{-1.6cm}
    \caption{A typical workflow for visual bug detection in video games.}
    \label{fig:bug_flowchart}
\end{figure*}

Using Computer Vision (CV) models~\cite{cv_game_test21} for automatic visual bug detection is a promising solution that can significantly reduce costs and manual labor. A typical workflow for utilizing a CV model for visual bug detection is depicted in Figure~\ref{fig:bug_flowchart}. The workflow starts with collecting labeled samples highlighting the bugs. Visual bugs are usually preferred to be indicated within a bounding box, making it easier to locate and fix the bugs. Consequently, an object detection annotation method~\cite{lin2014microsoft} is well-suited to meet this requirement. Afterwards, supervised model training is initiated on the collected datasets. During deployment time, the model predicts visual bugs on new deployment samples, followed by a triage stage to verify and isolate false-positive samples. Finally, the detected bugs are logged in the database.

Supervised machine learning-based visual bug detection is an increasingly promising approach for automating game testing workflows, offering effective and labor-efficient solutions~\cite{davarmanesh2020, taesiri2020, Azizi2023, Azizi2024, ling2020using}. Despite the effectiveness of the supervised learning methodology, its primary limitation is the requirement for a substantial volume of labeled training data. This is particularly challenging for multi-frame bugs, which typically have a low prevalence rate in video games. Current strategies to mitigate this issue include employing weakly supervised learning with synthetic data augmentations, which are generated using task-specific domain knowledge~\cite{rahman2023weak}. Another approach involves leveraging the inductive bias of pretrained multimodal large language models to reduce reliance on human supervision~\cite{taesirilarge}. In contrast to this prior work, our method is independent of task-specific domain knowledge and is computationally efficient, facilitating rapid deployment and broad accessibility for game development teams.

We propose a hybrid method utilizing both supervised and self-supervised learning objectives to leverage labeled data, providing a direct signal for bug detection, and unlabeled data, providing inductive bias on the visual semantics in the gaming domain. Moreover, our method could effectively fuse data from other games during training, which we refer to as co-finetuning, to further enhance the training data efficiency characteristics. During supervised training, we fuse the supervised loss from a target game with one from a mixture of other gaming titles using a linear gating scheme. While for self-supervised learning, we adopt a novel method following a Joint Embedding Predictive Architecture (JEPA)~\cite{ijepa23} that aims to reconstruct masked patches from unlabeled training samples on a target embedding space rather than the pixel space. Targeting reconstruction on the embedding space helps to focus the learned visual representation on high-level concepts while being robust to noise and irrelevant dynamics compared to working on the pixel space~\cite{jepa_approach_23}. Moreover, we aim to distill the target embedding from large pretrained vision encoders~\cite{kirillov2023segment,caron2021emerging,oquab2023dinov2}, which incorporates useful inductive bias and facilitates using efficient vision encoders during inference. 

To contrast the performance of our proposed method, we evaluate its bug detection performance using three different gaming environments, while comparing it to well-established object detection baselines~\cite{girshick2015fast,yolo16} from the Azure machine learning platform (Azure AutoML). We summarize our contribution points as follows:
\begin{itemize}
    \item We propose a co-finetuning method leveraging labeled data from a target game and a mixture of other games to maximize the utilization of data labeling efforts.
    \item We propose a novel self-supervised learning method that harnesses unlabeled gaming data to learn an effective visual representation and enhance data efficiency.
    \item We provide a comprehensive evaluation of the proposed method using multiple gaming environments resembling actual AAA games and comparing with well-established baselines trained and fine-tuned using Azure AutoML.
\end{itemize}

\section{RELATED WORK}
\label{sec:rw}
The emergence of machine learning has introduced automated and semi-automated approaches to enhance the efficiency of bug detection. Supervised methods for visual bug detection~\cite{davarmanesh2020, taesiri2020, Azizi2023, Azizi2024, ling2020using,tamm2022} are often limited by their reliance on large, labeled datasets, which are difficult to collect from specific game titles. To address this, some researchers have proposed weak supervision methods that incorporate synthetic data derived from task-specific domain knowledge to reduce the need for manual labeling~\cite{rahman2023weak}. Another approach, demonstrated in~\cite{arnab2022beyond}, shows that co-finetuning with multiple correlated supervised tasks, such as video action classification and object detection, can improve performance over traditional transfer learning. Our work builds upon this paradigm by combining self-supervised learning with supervised objectives. In a separate effort, a zero-shot video game bug detection method~\cite{taesirilarge} utilized inductive bias from a large multimodal language model to mitigate reliance on supervised training samples.

\section{PROBLEM DEFINITION}
\label{sec:pd}
Following an object detection objective, our primary aim is to accurately localize and classify visual bugs as objects within an image. Consequently, the problem is formulated as the minimization of a multi-task loss function, which balances these two essential components. We assume a region proposal approach~\cite{girshick2015fast} where the loss function comprises a weighted sum of the classification loss, $L_{cls}$, and the bounding box regression loss, $L_{loc}$, applied to both a Region Proposal Network (RPN) and the final detection head. Formally, for a given region proposal, the total loss is defined as:

\begin{equation}
\begin{split}
    L_{od} = \frac{1}{N_{cls}} \sum_i L_{cls}(p_i, p_i^*)  + \lambda \frac{1}{N_{loc}} \sum_i p_i^* L_{loc}(t_i, t_i^*)
\end{split}
\label{eq:od_supervised}
\end{equation}

Here, $i$ refers to the index of an anchor (for the RPN) or a region of interest (for the detection head). $p_i$ is the predicted probability of anchor $i$ being an object, while $p_i^*$ is its corresponding ground-truth label (1 for positive, 0 for negative). $t_i$ represents the predicted bounding box regression offsets for anchor $i$, and $t_i^*$ denotes the ground-truth bounding box regression offsets for a positive anchor $i$. $N_{cls}$ is the mini-batch size utilized for classification, and $N_{loc}$ is the number of anchor locations used for localization. Finally, $\lambda$ is a balancing weight used to normalize the contributions of the classification and localization losses.

\section{METHODOLOGY}
\label{sec:method}
\begin{figure}
    \centering
    \includegraphics[width=0.5\textwidth, height=4.5cm]{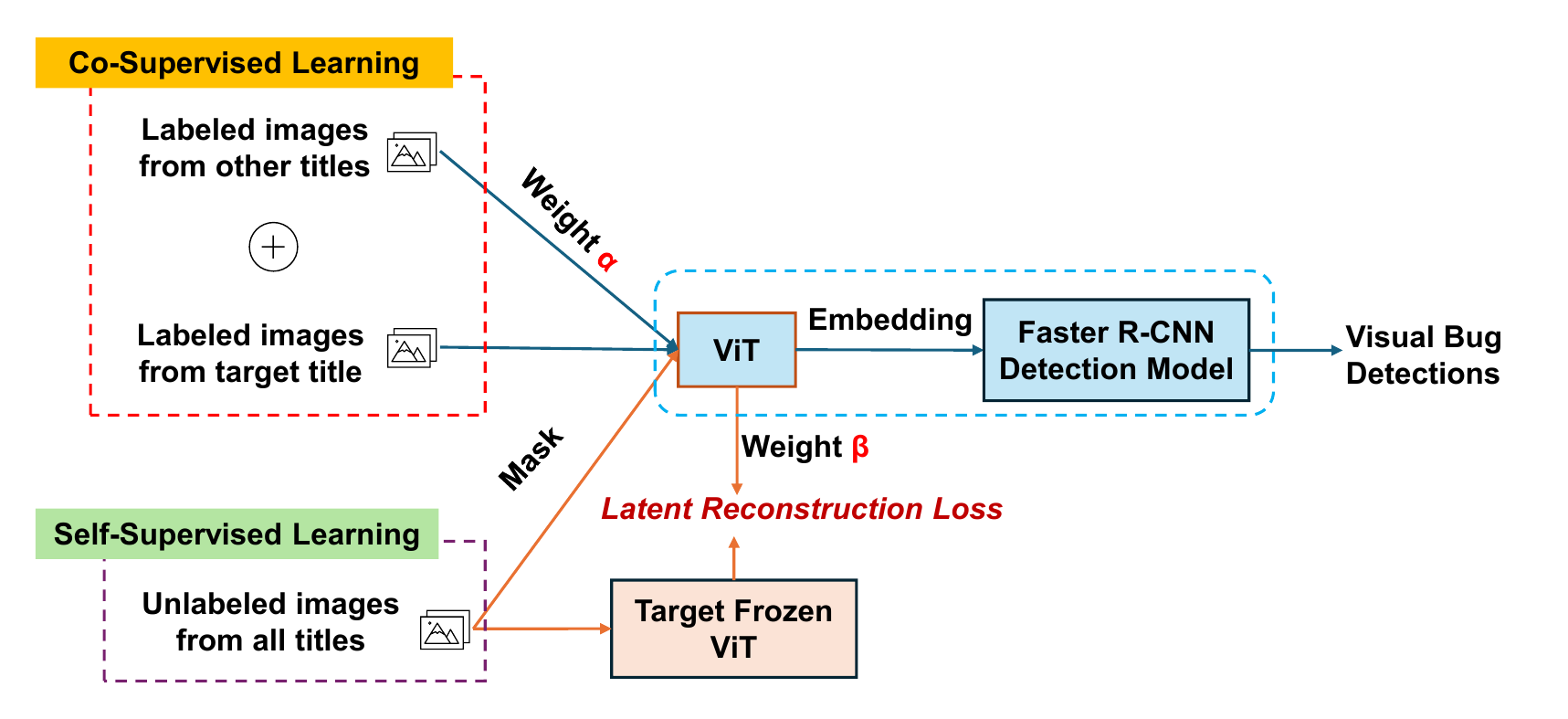}
    \caption{A block diagram for the Co-finetuning method design. It involves two main objectives: a self-supervised target distillation objective and a co-supervised object detection objective. }
    \label{fig:cft_design}
\end{figure}

In this section, we introduce the details of our proposed hybrid co-finetuning method. Figure~\ref{fig:cft_design} depicts the design of our proposed method, highlighting two main learning objectives, including: 1) co-supervised learning, and 2) self-supervised learning. This design aims to harness learning signals from both labeled and unlabeled data to enhance the detection performance. We outline each of these learning objectives as follows.

\subsection{Co-supervised Learning Module}
The objective of this learning module is to utilize labeled data from the downstream gaming title (i.e., the one under investigation) and from other gaming titles that might still provide relevant learning signals about the notion of bugs that we aim to detect. Thus, we combine labeled data into two groups, including the downstream and the co-title. During training, we sample mini-batches from each group and fuse the supervised loss (see Eq.~\ref{eq:od_supervised}) using a weight hyperparameter $\alpha$:
\begin{equation}
    L_{co\_sup} = L_{od}^{downstream} + \alpha\,L_{od}^{co-title}
\end{equation}

We adopt a Faster R-CNN ~\cite{girshick2015fast} object detection architecture with a Vision Transformer (ViT)~\cite{dosovitskiy2020image} backbone. The Faster R-CNN detection head includes a region proposal network, a classification head, and a box regression head. As per Equation~\ref{eq:od_supervised}, we utilize a cross-entropy loss for the $L_{cls}$ term, and a smooth L1 loss for the $L_{loc}$ term.

This co-finetuning algorithm includes both co-supervised learning and self-supervised learning components. For the co-supervised learning part, there are two types of labeled images: one from the target game title and the other from different game titles. Labeled images from multiple titles can increase the diversity of multi-frame bugs and enhance performance. The labeled images from the target downstream title and other titles should have different importance during the training phase. Therefore, we assign a weight $\alpha$ to the labeled images from other titles, which is reflected in the loss calculation.

\subsection{Self-supervised Learning Module}

\begin{figure}
    \centering
    \includegraphics[width=0.47\textwidth, height=5cm]{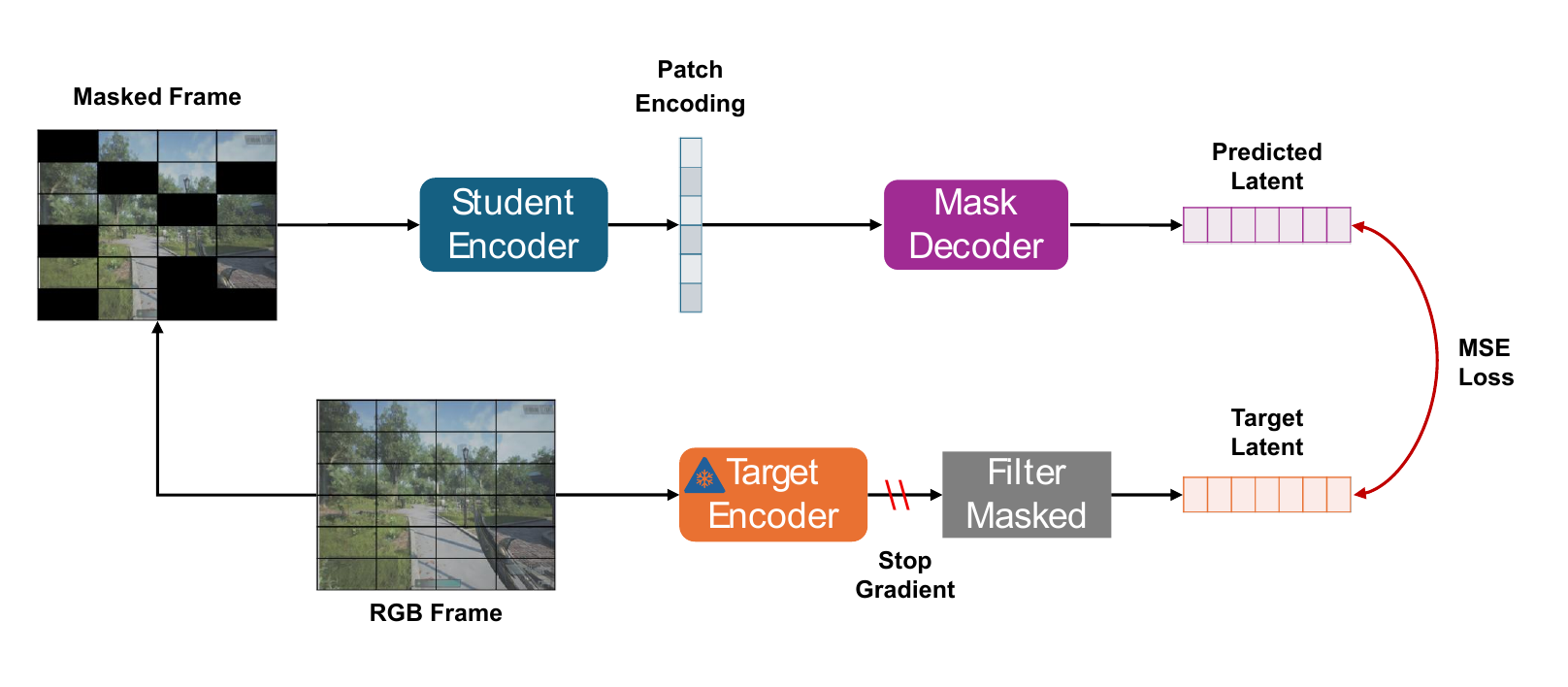}
    \caption{Illustration for our proposed self-supervised objective. We follow a latent masked autoencoder with a target distillation criterion, where an input frame is patchified and masked, then we feed the original version into a target encoder and the masked one to our student encoder, finally, we calculate the reconstruction error on the latent space, mainly considering masked patches.}
    \label{fig:ssl_design}
\end{figure}

To leverage additional learning signals from unlabeled data that could be collected at a cheaper cost in comparison to labeled the one, we propose a Self-supervised Learning (SSL) module that works as an auxiliary loss signal generator during training time. Our SSL design is inspired by the image Masked Autoencoders (MAE)~\cite{he2022masked} and Image-based Joint-Embedding Predictive Architecture (I-JEPA)~\cite{ijepa23}, in terms of using reconstruction after masking as a target and performing reconstruction on the latent space, respectively. Yet, our approach differs in using a separate target encoder from the one used by the architecture, which also acts as a distillation target during reconstruction. Figure~\ref{fig:ssl_design} illustrates the workflow of our SSL module. We combine images from all titles, including the downstream one, to form an unlabeled dataset. For a given unlabeled image sample, we divide it into equal-sized patches (similar to the ViT~\cite{dosovitskiy2020image} approach), then we create another replica by performing patch-wise random masking (with a masking ratio of 0.75) on it. We feed the original image to the target encoder, which is kept frozen during training, and the masked one to the student encoder used inside our architecture. Instead of calculating the reconstruction loss on the pixel space as in MAEs~\cite{he2022masked}, we target to reconstruct the masked patches on the latent space of the target encoder, similar to I-JEPA~\cite{ijepa23}. After encoding all the patches in the input image using the student encoder, including masked ones, we reconstruct the masked patches using a mask decoder. Finally, we match the reconstructed latents with the target ones from the target encoder to calculate our SSL loss. We note that we only focus our latent reconstruction loss on the masked patches without considering the unmasked ones by filtering them out of the loss computation. 

We utilize Mean Squared Error (MSE) to calculate the latent construction error, and we combine the SSL loss term with the co-supervised one using a hyperparameter $\beta$.
\begin{equation}
    L_{CFT} = L_{co\_sup} + \beta\,MSE_{ssl}
\end{equation}

\section{EXPERIMENTAL EVALUATION}
\label{sec:exp}

In this section, we introduce details of our experimental evaluation, including the dataset, baselines, experiments, and results.

\subsection{Data Collection and Splitting Strategy}

\begin{figure}[H]
    \centering
    \includegraphics[width=0.47\textwidth, height=3cm]{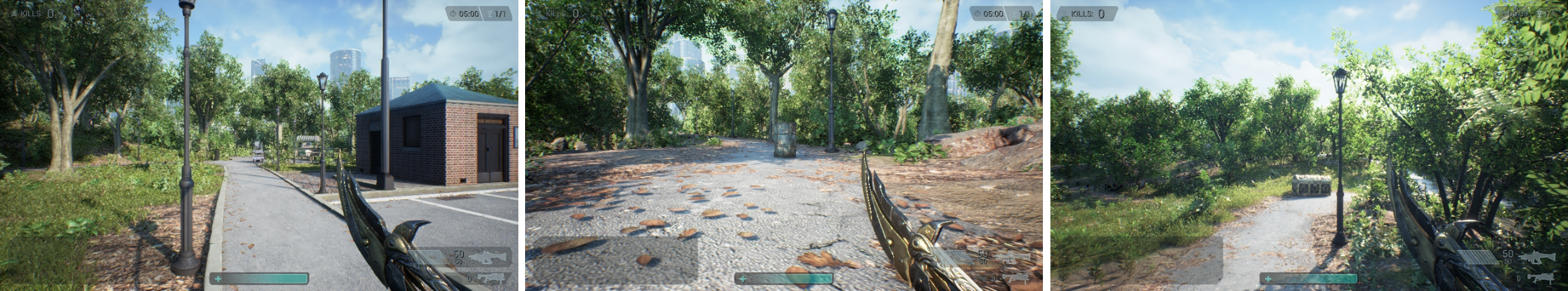}
    \caption{Representative frames from the GiantMap game.}
    \label{fig:giantmap_frame}
\end{figure}

\begin{figure}[H]
    \centering
    \includegraphics[width=0.47\textwidth, height=3cm]{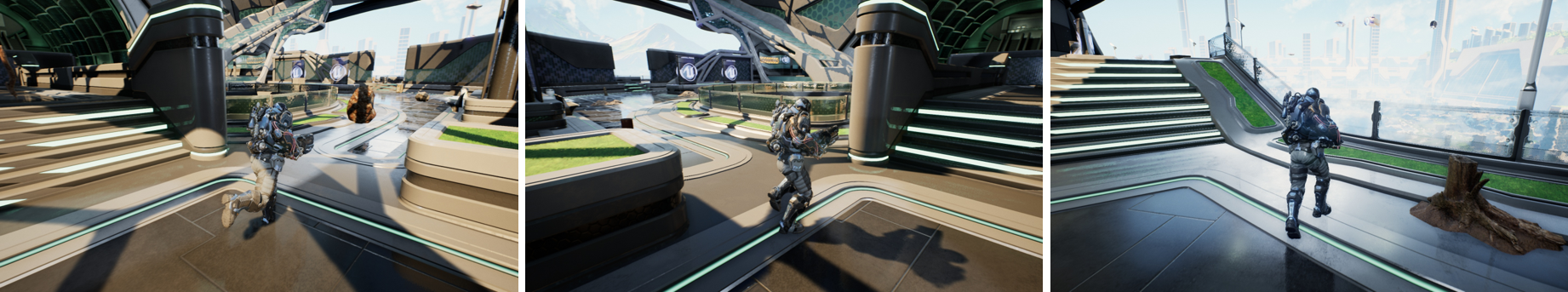}
    \caption{Representative frames from the HighRise game.}
    \label{fig:highrise_frame}
\end{figure}

We collected datasets from three distinct video game titles, each offering unique visual styles and gameplay dynamics. The selected titles include two experimental games that we developed using the Unreal Engine and one released Xbox game. We followed an automated data collection from experimental games using the UnrealCV plugin~\cite{unrealcv16} and manual collection from the released Xbox game. We outline the games as follows:
\begin{itemize}
  \item \textbf{GiantMap:} This game simulates a large city park area featuring open-world characteristics with trails, playgrounds, service facilities, and natural scenery. We scatter 3D assets around the park and randomly inject visual bugs into a certain portion of them. Representative frames from GiantMap are presented in Figure~\ref{fig:giantmap_frame}.
  
  \item \textbf{HighRise:} This game is part of the Shooter game demo from Unreal Engine. It simulates a futuristic indoor environment of a two-story building. HighRise is a fast-paced shooter that showcases advanced architectural designs and dynamic lighting effects. Representative frames from this game are illustrated in Figure~\ref{fig:highrise_frame}.

  \item \textbf{CombatGame:} This is a real combat-oriented AAA game that highlights tactical engagements, weapon-based combat, and rugged terrain. We do not show frames from this environment for privacy reasons.
\end{itemize}

\begin{figure*}
    \centering
    \includegraphics[scale=0.4]{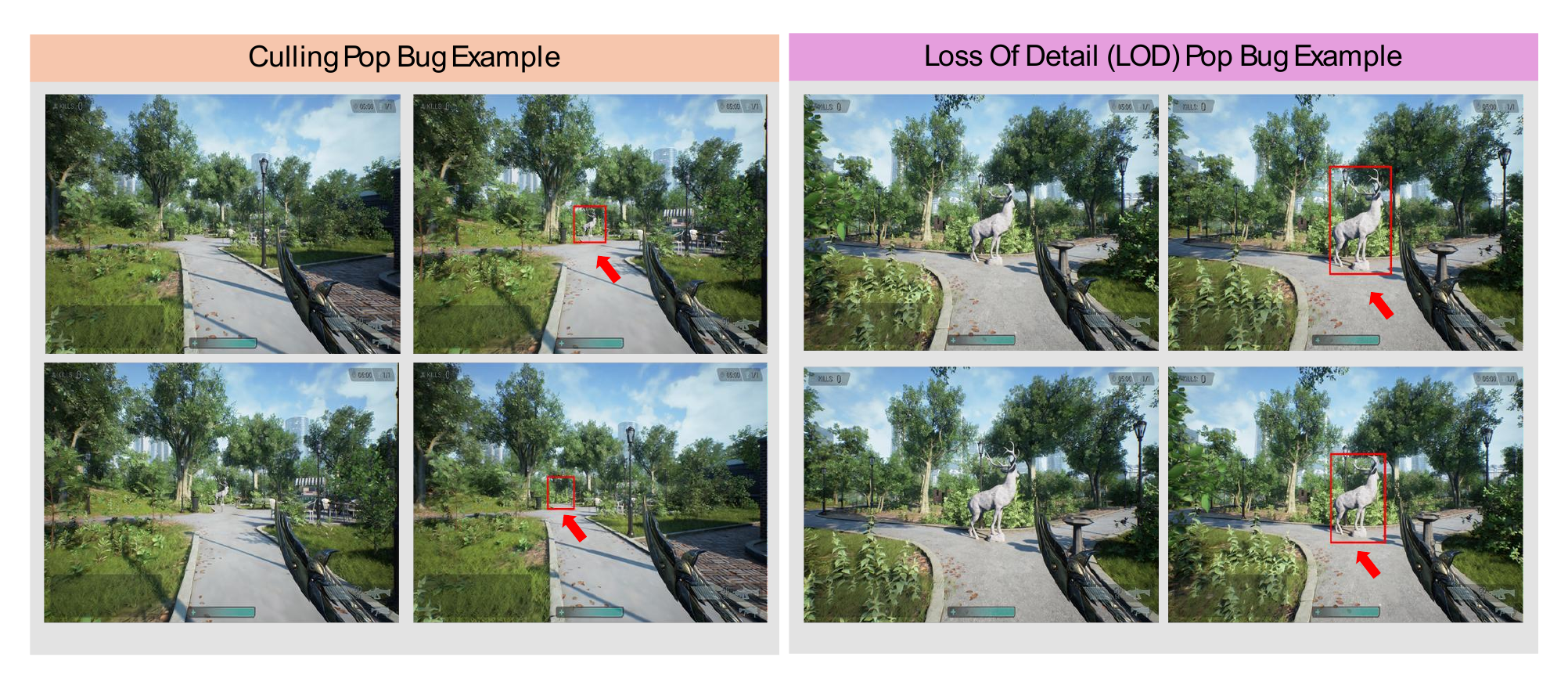}
    \caption{Two examples of pop visual bugs from the GiantMap game. On the left, a culling pop bug example. On the right, a LoD pop example. }
    \label{fig:pop_examples}
\end{figure*}

This study addresses multi-frame visual artifacts, specifically Level of Detail (LOD) pops and culling pops. An LOD pop occurs when an object's visual representation, such as its texture or geometric detail, undergoes an abrupt change as a viewer's distance to it varies, resulting in a discontinuous LOD transition. Conversely, a culling pop manifests as the sudden appearance or disappearance of an object within the rendered scene, typically in response to changes in viewer proximity or orientation. Illustrative examples of both culling and LOD pops are provided in Figure~\ref{fig:pop_examples}.

The dataset for each game title was partitioned into training, validation, and testing sets. The training set includes approximately 4,000 pops (GiantMap: 4503, CombatGame: 4661, Highrise: 2866), while both validation and testing sets each contain roughly 500 pops (Validation: GiantMap: 500, CombatGame: 543, Highrise: 351; Testing: GiantMap: 503, CombatGame: 507, Highrise: 352). Full details on pop and image counts for all four titles are in Table~\ref{tbl:data_splits}. Note that image counts may differ from pop counts, as one image can contain multiple pops. Each pop was annotated by one person, verified by a second, and adjudicated by a third in cases of disagreement. Additionally, 2,000 unlabeled images were randomly sampled per game title. For training on each target game, the other two games will serve as co-titles.

\begin{table*}[htbp]
\centering
\begin{tabular}{lccccccc}
\toprule
\multirow{2}{*}{Game Title} & \multicolumn{2}{c}{Training} & \multicolumn{2}{c}{Validation} & \multicolumn{2}{c}{Testing} & \multirow{2}{*}{\makecell{Number of\\Unlabeled Images}} \\
\cmidrule(lr){2-3} \cmidrule(lr){4-5} \cmidrule(lr){6-7}
 & \# of pop & \# of image & \# of pop & \# of image & \# of pop & \# of image & \\
\midrule
GiantMap & 4503 & 4491 & 500 & 500 & 503 & 499 & 2000 \\
CombatGame     & 4661 & 1574 & 547 & 146 & 507 & 247 & 2000 \\
Highrise & 2866 & 2856 & 351 & 347 & 352 & 345 & 2000 \\
\bottomrule
\end{tabular}
\caption{Train-Validation-Test data splits and number of unlabeled samples across the experimental gaming datasets.}
\label{tbl:data_splits}
\end{table*}

\subsection{Comparison Baselines}
 We use the object detection models provided by Azure AutoML as our baseline. Azure AutoML is a cloud-based service within Microsoft Azure’s Automated Machine Learning suite that enables users to automatically build, train, and deploy machine learning models, including those for object detection. Azure AutoML supports two types of object detection architectures: a two-stage Faster R-CNN detector~\cite{girshick2015fast} and a single-stage YOLOv5 detector~\cite{yolo16, jocher2020ultralytics}. When running object detection tasks, Azure AutoML automatically explores multiple variants of these models and selects the best-performing model based on a validation dataset and a specified target metric. We refer to the selected Azure AutoML variant as ``AutoML'' in our result reporting for brevity. In our case, the target metric is mean average precision (mAP)~\cite{girshick2015fast}. Thus, our comparison baseline is not a single fixed model, but rather the result of an automated search across state-of-the-art deep learning architectures. All hyperparameters are tuned using the validation data, and the final model is chosen based on its performance against the mAP metric.

\begin{figure*}
    \centering
    \includegraphics[scale=0.5]{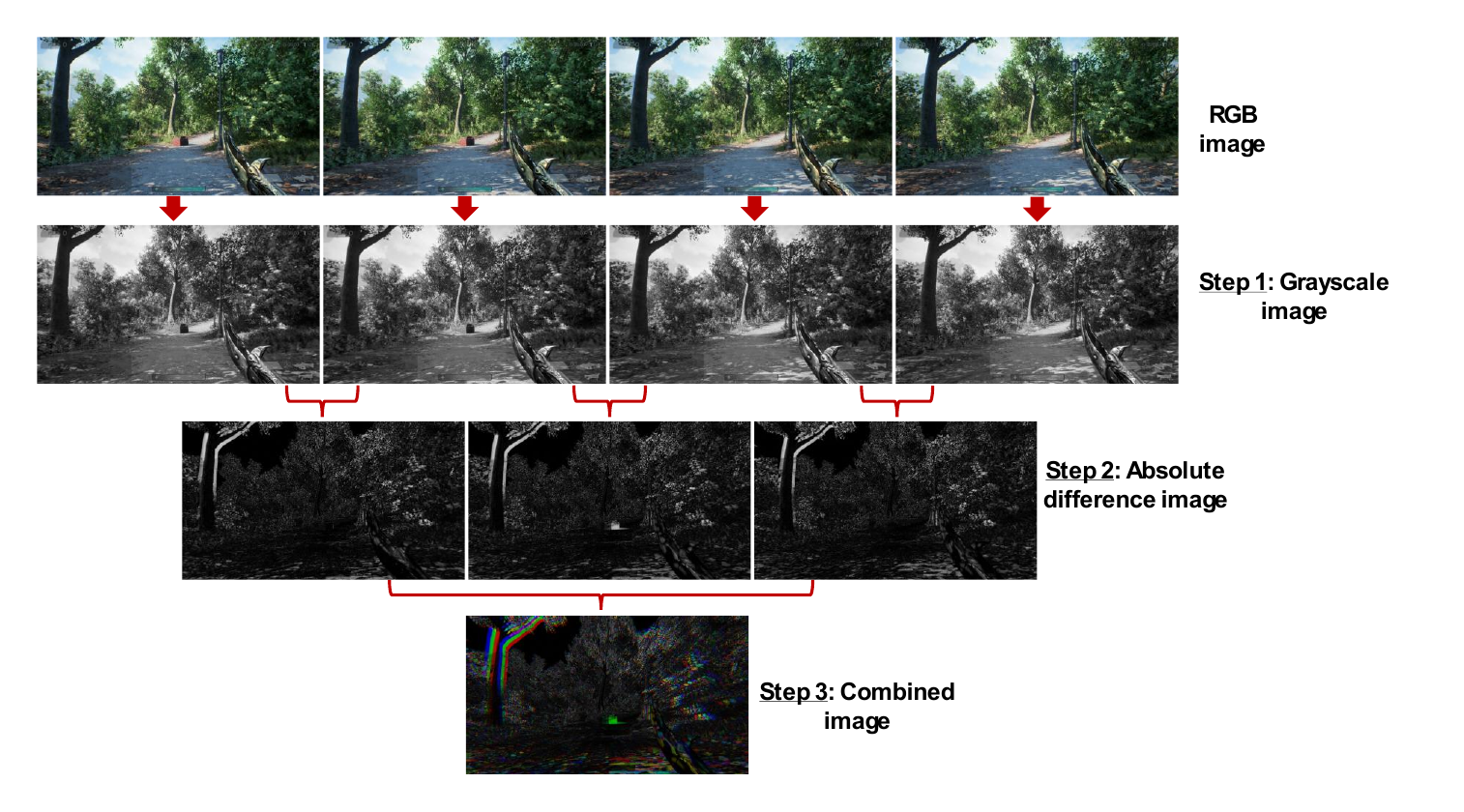}
    \caption{An illustration of the input sliding window preparation procedure. RGB images are first converted to grayscale, then absolute deltas are computed, and finally, these deltas are combined into a three-channel image.}
    \label{fig:img_preparation}
\end{figure*}

For multi-frame pop data, the pop bug only becomes apparent when analyzing sequences of frames. Therefore, using a single image as input is insufficient for detecting the issue. However, Azure AutoML object detection models require input images to have exactly three channels. To address this, we construct a three-channel image using a grayscale difference stacking approach. The idea is to use a sequence of four RGB images, where the pop bug occurs in the third frame. These four RGB images are first converted to grayscale. Then, we compute the absolute differences between consecutive grayscale frames to capture motion or changes across time. These different images are stacked to form a new three-channel image suitable for input into the Azure AutoML model.The process is further described as follows: four RGB frames-\textit{rgb}$_0$, \textit{rgb}$_1$, \textit{rgb}$_2$, and \textit{rgb}$_3$-are used, with the pop bug occurring in \textit{rgb}$_2$. Each RGB image is converted to grayscale, resulting in \textit{gray}$_0$, \textit{gray}$_1$, \textit{gray}$_2$, and \textit{gray}$_3$. Then, the absolute differences between consecutive grayscale images are computed: $ \mathit{diff}_1 = \left| \mathit{gray}_1 - \mathit{gray}_0 \right| $, $ \mathit{diff}_2 = \left| \mathit{gray}_2 - \mathit{gray}_1 \right| $, and $ \mathit{diff}_3 = \left| \mathit{gray}_3 - \mathit{gray}_2 \right| $ where $|\cdot|$  denotes the absolute value operation. These three different images are stacked to form a new three-channel image, denoted as img, which is subsequently used as input for both the Azure AutoML object detection model and our proposed CFT model. This image preparation step for the model input is also illustrated in Figure~\ref{fig:img_preparation}.

\subsection{Research Questions}
Our experimental evaluation setup aims to assess the different aspects of the proposed hybrid co-finetuning method by answering the following research questions. 

\begin{itemize}
    \item \textbf{[RQ1]}: How does the visual bug detection performance of the proposed method compare to well-established object detection baselines using the evaluation gaming environments?
    \item \textbf{[RQ2]}: What would be the suitable mixing strategy when learning from mixed gaming data?
    \item \textbf{[RQ3]}: What is the impact of different building blocks in the proposed hybrid co-finetuning method?
    \item  \textbf{[RQ4]}: Could the proposed hybrid co-finetuning method generalize its performance to other types of visual bugs in games?
\end{itemize} 

\subsection{Results \& Discussion}

In this section, we present and discuss the evaluation results answering the defined research questions.

\subsubsection{Visual Bug Detection Performance Comparison with Baselines: }We answer \textbf{RQ1} by assessing the CFT model performance compared to Azure AutoML baselines in terms of key evaluation metrics—mean Average Precision (mAP)~\cite{EveringhamGWWZ10} and F1 score ~\cite{goutte2005probabilistic}. A model is considered effective if it achieves higher mAP and F1 scores than the baseline. However, even if the CFT model delivers comparable performance, it may still be preferable if it requires significantly less labeled data. This consideration is particularly important in the gaming domain, where acquiring labeled bug data is both time-consuming and resource-intensive. Therefore, a model that maintains competitive performance while reducing the need for labeled data offers a substantial practical advantage.

 Azure AutoML automatically evaluates both Faster R-CNN and YOLO V5 for bug detection and selects the best performing baseline based on the validation dataset. We use mean average precision (mAP) with an intersection over union (IoU) threshold of 0.5 for pop bug detection. Additionally, we measure the F1 score with IoU=0.5 and a classification threshold of 0.25. We select a threshold of 0.25 for result comparison because recall is usually more important for visual bug classification. The results for these metrics, based on the testing dataset for Azure AutoML and the co-finetuning (CFT) method, are presented in Table~\ref{tbl:automl_cft}.

\begin{table*}[htbp]
\centering
\begin{tabular}{lcccccccc}
\toprule
\multirow{3}{*}{Dataset} & \multicolumn{4}{c}{mAP} & \multicolumn{4}{c}{F1} \\
\cmidrule(lr){2-5} \cmidrule(lr){6-9}
 & \multicolumn{2}{c}{AutoML} & \multicolumn{2}{c}{CFT} & \multicolumn{2}{c}{AutoML} & \multicolumn{2}{c}{CFT} \\
 & 100\% & 50\% & 100\% & 50\% & 100\% & 50\% & 100\% & 50\% \\
\midrule

Giantmap & 0.662 & 0.560 & \textbf{0.938} & \textbf{0.945} & 0.747 & 0.777 & \textbf{0.976} & \textbf{0.969} \\
CombatGame     & 0.159 & 0.106 & \textbf{0.164} & \textbf{0.158} & 0.418 & 0.337 & \textbf{0.445} & \textbf{0.450} \\
Highrise & 0.209 & 0.202 & \textbf{0.349} & \textbf{0.235} & 0.312 & 0.355 & \textbf{0.533} & \textbf{0.455} \\
\bottomrule
\end{tabular}
\caption{Visual bug detection performance results compared to the best performing Azure AutoML baseline using full and half the size of the training datasets.}
\label{tbl:automl_cft}
\end{table*}

In order to perform a statistical significance t-test for mAP and F1 metrics for each pair, we normalized each metric value for each game title. For example, we normalized the mAP values $[0.662, 0.938]$ from different algorithms to $[0.413=0.662/(0.662+0.938), 0.586=0.938/(0.662+0.938)]$ for game title GiantMap. This normalization is necessary because each game title has its own testing dataset, resulting in varied scales that make cross-title comparisons incomparable. The results are shown in Table~\ref{tbl:automl_cft_ttest}. Here, three samples were used, with each title contributing a single value to the statistical tests. It can be seen from  Table~\ref{tbl:automl_cft_ttest} that the difference between the results from Azure AutoML and CFT is statistically significant at the 0.05 level, even with $50\%$ training data in CFT, indicating that CFT achieves better results than Azure AutoML.
Furthermore, the difference between CFT using the full training data and $50\%$ of the training data is not statistically significant, suggesting that even with less training data, CFT can still produce comparable results. Additional t-tests were conducted on these results, as shown in Table~\ref{tbl:automl_cft_ttest}. It can be observed that reducing the training dataset does not significantly degrade performance in most cases for CFT in terms of mAP. However, when the training dataset is halved in Azure AutoML, the performance drops considerably. In contrast, CFT maintains better performance in terms of mAP metrics compared to the Azure AutoML baseline when trained on only half of the dataset.

\begin{table}[htbp]
\centering
\begin{tabular}{llr}
\toprule
Metric & T-test & P-value \\
\midrule
\multirow{4}{*}{mAP} 
  & AutoML (100\%) vs CFT (100\%) & 0.007 \\
  & AutoML (100\%) vs CFT (50\%) & 0.026 \\
  & CFT (100\%) vs CFT (50\%)    & 0.054 \\
  & AutoML (50\%) vs CFT (50\%) & 0.009 \\
  & AutoML (100\%) vs AutoML (50\%) & 0.007 \\
\midrule
\multirow{4}{*}{F1} 
  & AutoML (100\%) vs CFT (100\%) & 0.004 \\
  & AutoML (100\%) vs CFT (50\%) & 0.001 \\
  & CFT (100\%) vs CFT (50\%)    & 0.079 \\
  & AutoML (50\%) vs CFT (50\%) & 0.000 \\
  & AutoML (100\%) vs AutoML (50\%) & 0.965 \\
\bottomrule
\end{tabular}
\caption{Statistical significance t-test analysis comparing the CFT method with the best performing baseline selected by Azure AutoML using full and half the size of the training data.}
\label{tbl:automl_cft_ttest}
\end{table}

\subsubsection{Assessing the Data Mixing Strategy for Co-Finetuning: }The CFT model incorporates both co-supervised and self-supervised learning. However, optimizing the balance of these components, specifically their weighting during training, remains an open question as indicated in \textbf{RQ2}. This study investigates the impact of explicitly tuning this mixing strategy on model performance. An unsuitable weighting scheme could either underutilize labeled data or fail to fully leverage self-supervised learning benefits. Through simulations, we aim to understand the role of this mixing strategy, determining whether a fixed or adaptive weighting approach yields superior results and if model performance is sensitive to these choices. This investigation will clarify whether the mixing strategy should be considered a critical hyperparameter for the CFT model. We kept all other algorithmic parameters fixed and systematically varied the weights assigned to the co-supervised learning (CSL) and self-supervised learning (SSL) components, denoted as $\alpha$ and $\beta$, respectively (see Figure~\ref{fig:cft_design}). To determine the optimal values for these hyperparameters, we conducted a grid search over a predefined range of $\alpha$ and $\beta$, spanning from 0 to 0.6 in increments of 0.1. The best combination was selected based on performance on the validation datasets across three distinct game titles.

Table~\ref{tbl:csl_ssl_weights} summarizes the results, indicating that the optimal $\beta$ value remains relatively consistent across titles, centering around 0.2. Conversely, $\alpha$ demonstrates greater variability, suggesting its optimal value is more sensitive to individual game title characteristics. The magnitude of $\alpha$ generally exceeds $\beta$, implying the CSL component holds a more influential role in bug prediction performance. These findings suggest that while $\beta$ can be fixed at 0.2 for new titles, careful tuning of $\alpha$ is recommended for dataset adaptation. This insight can streamline hyperparameter optimization when extending the visual bug detection framework to new game titles.

\begin{table}[H]
\centering
\begin{tabular}{lcccc}
\toprule
Title & mAP & F1 & $\beta$ & $\alpha$ \\
\midrule
GiantMap & 0.941 & 0.890 & 0.3 & 0.4 \\
\midrule
CombatGame & 0.164 & 0.404 & 0.2 & 0.5 \\
\midrule
Highrise & 0.384 & 0.471 & 0.2 & 0.3 \\
\bottomrule
\end{tabular}
\caption{Optimal weighting schemes for CSL and SSL across experimental gaming environments.}
\label{tbl:csl_ssl_weights}
\end{table}

\subsubsection{Ablation Study: }To address \textbf{RQ3}, which aims to assess the impact of the CFT model's constituent components, we conduct an ablation study evaluating the necessity and contribution of its core elements: co-supervised learning (CSL) and self-supervised learning (SSL). This involves incrementally adding CSL and SSL components to determine their influence on performance. Additionally, we investigate the effect of different backbone architectures, specifically ViT-Base~\cite{dosovitskiy2020image} and ResNet-50~\cite{he2016deep}, to assess the model's sensitivity to backbone choice and identify optimal architectures for the given tasks. Finally, we explore the impact of various target encoders within the SSL component, evaluating pretrained encoders from state-of-the-art self-supervised learning methods such as DINOv1~\cite{caron2021emerging}, DINOv2~\cite{oquab2023dinov2}, MAE~\cite{he2022masked}, and SAM~\cite{kirillov2023segment}, all based on the ViT-Base architecture. This analysis will clarify how target encoder selection influences the quality of learned representations and overall effectiveness.

\begin{figure}[H]
    \centering
    \includegraphics[width=0.45\textwidth]{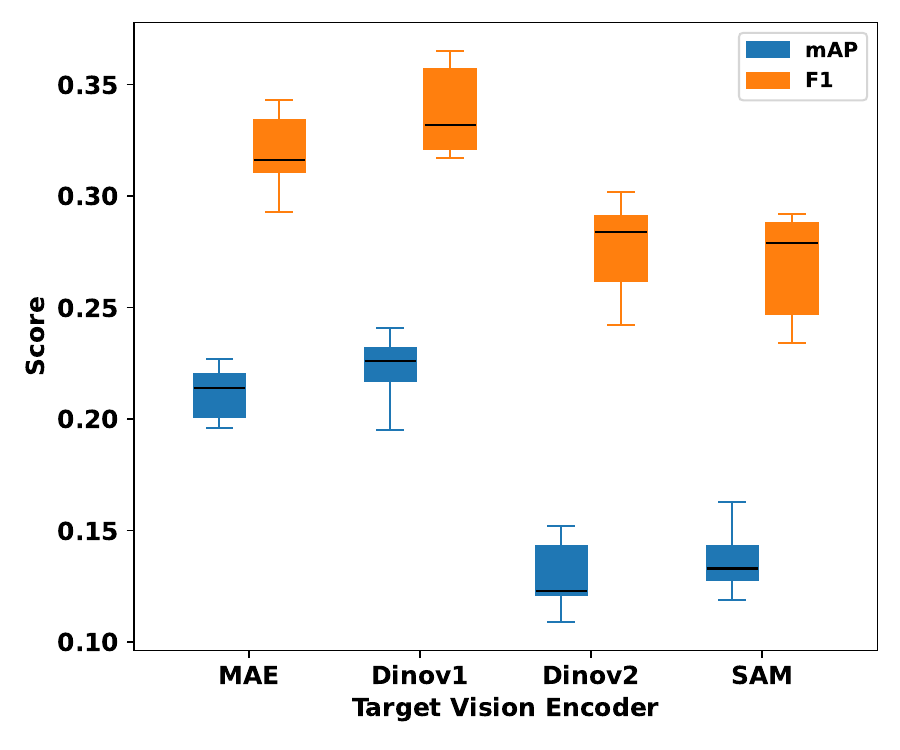}
    \caption{Assessing the impact of using different target vision encoders on the CFT method performance.}
    \label{fig:target_encoder}
\end{figure}

We also experimented with the widely adopted ResNet-50~\cite{he2016deep} backbone. It is crucial to note the differing input image dimensions: $600\times600$ pixels for ResNet-50 versus $224\times224$ pixels for ViT. The corresponding performance metrics, including mean Average Precision (mAP) and F1 score, are presented in Table~\ref{tbl:csl_ssl_ablation}. A comparison of these results with those obtained using the ViT-base backbone (Table~\ref{tbl:automl_cft}) reveals relatively minor differences in mAP and F1 score. However, the metrics for each game title appear slightly better when using the ViT-base backbone compared to ResNet-50. To rigorously evaluate this observation, we conducted a paired t-test on the mAP scores obtained from both backbones. The resulting p-value of 0.041 indicates that the performance difference is statistically significant at the conventional p-value=0.05 threshold. This finding suggests that the ViT-base backbone yields superior visual bug detection performance relative to ResNet-50, reinforcing its potential as a more effective architecture for this task.

To evaluate the individual contributions of CSL and SSL to visual bug detection, we conducted an ablation study using a ResNet-50 backbone. We assessed the CFT model's performance across four configurations: with both CSL and SSL enabled, with only CSL, with only SSL, and with neither.

Table~\ref{tbl:csl_ssl_ablation} summarizes these results, where "True" denotes inclusion and "False" denotes exclusion of a component. The highest performance, in terms of both mean Average Precision (mAP) and F1 score, is observed when both CSL and SSL are enabled. Conversely, the absence of both components yields the lowest performance, underscoring their collective importance.

Interestingly, enabling only SSL does not significantly improve performance over the baseline (neither component enabled), suggesting that SSL alone may be insufficient in this context. However, the inclusion of CSL, even without SSL, leads to substantial performance gains, indicating CSL's more dominant role in enhancing the model's comprehension of visual bugs. Nevertheless, SSL provides complementary benefits, as its addition to a CSL-enabled model further boosts performance. This demonstrates that while CSL is highly impactful independently, SSL contributes additional value synergistically.

\begin{table}[H]
\centering
\begin{tabular}{lcccc}
\toprule
 & P-Value(mAP) & P-Value(F1) \\
 \midrule
MAE vs SAM & 0.0 & 0.0077 \\
\midrule
DINOv2 vs SAM & 0.5036 & 0.6184 \\
\midrule
DINOv1 vs MAE & 0.3084 & 0.1834 \\
\midrule
DINOv2 vs MAE & 0.0 & 0.0145 \\
\midrule
DINOv1 vs DINOv2 & 0.0 & 0.0025 \\
\bottomrule
\end{tabular}
\caption{Statistical significance t-test analysis for using different target vision encoders in the self-supervised learning objective of the CFT method.}
\label{tbl:target_encoder_ttest}
\end{table}

We further investigate the impact of different target vision encoders used in the SSL component of the CFT model. Specifically, we evaluate four widely adopted pretrained Vision Transformer (ViT) encoders: DINOv1, DINOv2, MAE, and SAM. Each vision encoder is seamlessly integrated into the CFT framework. To ensure statistical robustness, we perform five independent simulation runs for each configuration using the half Highrise training dataset, which accelerates simulation without compromising generality. The results are visualized in the box-and-whisker plots shown in Figure~\ref{fig:target_encoder}. From Figure~\ref{fig:target_encoder}, we observe that the MAE and DINOv1 vision encoders yield highly overlapping distributions in both mAP and F1 score, suggesting comparable performance. To statistically validate this observation, we perform a two-sample t-test between the results of DINOv1 and MAE mAP. The resulting p-value of 0.3084 indicates that the performance difference between these two vision encoders is not statistically significant at the 0.05 level. In contrast, the performance of DINOv2 and SAM encoders is notably lower. A t-test comparing MAE and SAM mAP yields a p-value of ~0.0, confirming a statistically significant difference in performance. More t-test results are given in Table~\ref{tbl:target_encoder_ttest}. These findings suggest that while DINOv2 and SAM may be effective in other contexts, they are less suitable as target encoders in the SSL component of the CFT model. These results indicate that both MAE and DINOv1 serve as strong candidates for the target encoder in SSL, offering significantly better performance than DINOv2 and SAM. Given their similar effectiveness, either MAE or DINOv1 can be confidently selected for use in the CFT framework.

\begin{table*}[ht!]
\centering
\begin{tabular}{lcccccccc}
\toprule
\multirow{2}{*}{} & \multicolumn{2}{c}{SSL: True, CSL: True} & \multicolumn{2}{c}{SSL: True, CSL: False} & \multicolumn{2}{c}{SSL: False, CSL: True} & \multicolumn{2}{c}{SSL: False, CSL: False} \\
\cmidrule(lr){2-3} \cmidrule(lr){4-5} \cmidrule(lr){6-7} \cmidrule(lr){8-9}
 & mAP & F1 & mAP & F1 & mAP & F1 & mAP & F1 \\
\midrule
GiantMap & \textbf{0.936} & \textbf{0.918} & 0.901 & 0.842 & 0.918 & 0.890 & 0.898 & 0.836 \\
CombatGame     & \textbf{0.154} & \textbf{0.438} & 0.062 & 0.295 & 0.144 & 0.413 & 0.056 & 0.281 \\
Highrise & \textbf{0.330} & \textbf{0.434} & 0.061 & 0.169 & 0.329 & 0.418 & 0.057 & 0.090 \\
\bottomrule
\end{tabular}
\caption{Performance impact of co-supervised learning (CSL) and self-supervised (SSL) objectives on the CFT model performance.}
\label{tbl:csl_ssl_ablation}
\end{table*}

\subsubsection{Generalization to Other Visual Bug Types in Games: } Answering research question \textbf{RQ4}, we evaluate the proposed method's generalization capabilities for detecting single-frame visual bugs. These bugs primarily include floating object artifacts and texture distortions. Floating object bugs manifest when objects, typically expected to rest on a surface, appear suspended above it, creating a visible gap. Texture bugs are categorized as low-resolution texture defects, characterized by blurriness or lack of detail, and stretched texture distortions, where textures appear unnaturally elongated or distorted on an object's surface.

We evaluated the CFT model on these single-frame bug categories using the GiantMap and HighRise datasets. The GiantMap dataset includes 1,320 training, 120 validation, and 960 testing samples. The HighRise dataset comprises 3,326 training, 287 validation, and 2,450 testing samples. Evaluation results, measured by mean Average Precision (mAP) on the test sets, are presented in Table~\ref{tbl:automl_cft_single_frame}. As shown, CFT consistently outperforms AutoML on the GiantMap title for both floating and texture single-frame bug categories, regardless of the training dataset size (full or half). This highlights CFT's robustness and effectiveness in this context. Conversely, for the HighRise title, CFT and AutoML exhibit comparable performance, achieving similar results, particularly within the texture bug category.

\begin{table*}[htbp]
\centering
\begin{tabular}{lcccccccc}
\toprule
\multirow{3}{*}{Game Title} & \multicolumn{4}{c}{Floating} & \multicolumn{4}{c}{Texture} \\
\cmidrule(lr){2-5} \cmidrule(lr){6-9}
 & \multicolumn{2}{c}{AutoML} & \multicolumn{2}{c}{CFT} & \multicolumn{2}{c}{AutoML} & \multicolumn{2}{c}{CFT} \\
 & 100\% & 50\% & 100\% & 50\% & 100\% & 50\% & 100\% & 50\% \\
\midrule

Giantmap & 0.831 & 0.811 & \textbf{0.864} & \textbf{0.850} & 0.796 & 0.781 & \textbf{0.828} & \textbf{0.803} \\
Highrise & \textbf{0.733} & \textbf{0.690} & 0.721 & 0.618 & 0.965 & 0.954 & \textbf{0.968} & \textbf{0.956} \\
\bottomrule
\end{tabular}
\caption{Single frame visual bug detection performance results comparing the CFT method with the best performing Azure AutoML baseline.}
\label{tbl:automl_cft_single_frame}
\end{table*}

\section{IMPLEMENTATION  DETAIL}
Model training was primarily conducted on a server featuring an NVIDIA RTX A6000 GPU with 49 GB of memory, running Ubuntu 18.04. Training spanned 30 epochs with a base learning rate of 0.0001. The per-device batch size was initialized to 10. To achieve a larger effective batch size without exceeding GPU memory, we utilized gradient accumulation with an accumulation step of 10, resulting in an effective batch size of 100 (10×10=100). When adapting to different hardware, the per-device batch size was adjusted, while maintaining an effective batch size of 100 by proportionally modifying the accumulation step, ensuring consistent training dynamics. For optimization, the AdamW optimizer~\cite{loshchilov2017decoupled} was employed alongside a cosine annealing learning rate schedule~\cite{loshchilov2016sgdr}. A warm-up phase was applied during the initial 10 epochs to stabilize early training. Additionally, L2 regularization with a weight decay coefficient of 0.0005 was used to mitigate overfitting and enhance generalization.

\section{CONCLUSION}
\label{sec:conclusion}
We introduce a hybrid co-finetuning (CFT) method for multi-frame visual bug detection, which enhances performance and data efficiency by leveraging both labeled and unlabeled training data. This is achieved through a hybrid workflow that fuses supervised and self-supervised learning signals from the target game domain with a co-domain derived from a mixture of other games.

Our comprehensive evaluation compared the proposed method against established baselines across three distinct AAA game titles. The performance results indicate that our method consistently outperforms the best baseline on all datasets. Notably, it maintains a statistically significant performance advantage even when trained with half the size of the original datasets. We also conducted an extensive ablation study to evaluate the impact of different architectural components within the proposed method. Finally, we demonstrated the method's generalization potential by applying it to other types of visual bugs sampled from two additional gaming environments.

Our future work directions are outlined. We aim to investigate methodologies for learning a shared embedding space across multiple vision modalities~\cite{BachmannMAZ22,GirdharSRMJM22} during training time while restricting inference to the pixel modality for runtime efficiency. Another direction involves exploring adaptive data mixing strategies~\cite{HejnaBJPS24} to identify an optimal scheme tailored to specific downstream gaming domains. Finally, we plan to cover additional multi-frame visual bug types, such as Z-fighting or lighting bugs. In terms of training time, under identical hyperparameter settings, the baseline model for Giantmap trains in about 1.5 hours, while the CFT takes roughly 9.3 hours. Accelerating CFT training remains a direction for further study.

In conclusion, our CFT method effectively demonstrates the benefits of supervised co-finetuning using a mixture of other games, while simultaneously leveraging unlabeled data through self-supervised learning as an auxiliary objective.

\bibliography{references}

\begin{thebibliography}{32}
\providecommand{\natexlab}[1]{#1}
\providecommand{\url}[1]{\texttt{#1}}
\expandafter\ifx\csname urlstyle\endcsname\relax
  \providecommand{\doi}[1]{doi: #1}\else
  \providecommand{\doi}{doi: \begingroup \urlstyle{rm}\Url}\fi

\bibitem[Arnab et~al.(2022)Arnab, Xiong, Gritsenko, Romijnders, Djolonga,
  Dehghani, Sun, Lu{\v{c}}i{\'c}, and Schmid]{arnab2022beyond}
Arnab, A., Xiong, X., Gritsenko, A., Romijnders, R., Djolonga, J., Dehghani,
  M., Sun, C., Lu{\v{c}}i{\'c}, M., and Schmid, C.
\newblock Beyond transfer learning: Co-finetuning for action localisation.
\newblock \emph{arXiv preprint arXiv:2207.03807}, 2022.

\bibitem[Assran et~al.(2023)Assran, Duval, Misra, Bojanowski, Vincent, Rabbat,
  LeCun, and Ballas]{ijepa23}
Assran, M., Duval, Q., Misra, I., Bojanowski, P., Vincent, P., Rabbat, M.~G.,
  LeCun, Y., and Ballas, N.
\newblock Self-supervised learning from images with a joint-embedding
  predictive architecture.
\newblock In \emph{{IEEE/CVF} Conference on Computer Vision and Pattern
  Recognition, {CVPR} 2023, Vancouver, BC, Canada, June 17-24, 2023}, pp.\
  15619--15629. {IEEE}, 2023.

\bibitem[Azizi \& Zaman(2023)Azizi and Zaman]{Azizi2023}
Azizi, E. and Zaman, L.
\newblock Automatic bug detection in games using lstm networks.
\newblock In \emph{2023 IEEE Conference on Games (CoG)}, pp.\  1--4, 2023.

\bibitem[Azizi \& Zaman(2024)Azizi and Zaman]{Azizi2024}
Azizi, E. and Zaman, L.
\newblock Astrobug: Automatic game bug detection using deep learning.
\newblock \emph{IEEE Transactions on Games}, 16\penalty0 (4):\penalty0
  793--806, 2024.

\bibitem[Bachmann et~al.(2022)Bachmann, Mizrahi, Atanov, and
  Zamir]{BachmannMAZ22}
Bachmann, R., Mizrahi, D., Atanov, A., and Zamir, A.
\newblock Multimae: Multi-modal multi-task masked autoencoders.
\newblock In Avidan, S., Brostow, G.~J., Ciss{\'{e}}, M., Farinella, G.~M., and
  Hassner, T. (eds.), \emph{Computer Vision - {ECCV} 2022 - 17th European
  Conference, Tel Aviv, Israel, October 23-27, 2022, Proceedings, Part
  {XXXVII}}, volume 13697 of \emph{Lecture Notes in Computer Science}, pp.\
  348--367. Springer, 2022.

\bibitem[Caron et~al.(2021)Caron, Touvron, Misra, J{\'e}gou, Mairal,
  Bojanowski, and Joulin]{caron2021emerging}
Caron, M., Touvron, H., Misra, I., J{\'e}gou, H., Mairal, J., Bojanowski, P.,
  and Joulin, A.
\newblock Emerging properties in self-supervised vision transformers.
\newblock In \emph{Proceedings of the IEEE/CVF international conference on
  computer vision}, pp.\  9650--9660, 2021.

\bibitem[Davarmanesh et~al.(2020)Davarmanesh, Jiang, Ou, Vysogorets,
  Ivashkevich, Kiehn, Joshi, and Malaya]{davarmanesh2020}
Davarmanesh, P., Jiang, K., Ou, T., Vysogorets, A., Ivashkevich, S., Kiehn, M.,
  Joshi, S.~H., and Malaya, N.
\newblock Automating artifact detection in video games.
\newblock \emph{arXiv preprint arXiv:2011.15103}, 2020.

\bibitem[Dawid \& LeCun(2023)Dawid and LeCun]{jepa_approach_23}
Dawid, A. and LeCun, Y.
\newblock Introduction to latent variable energy-based models: {A} path towards
  autonomous machine intelligence.
\newblock \emph{CoRR}, abs/2306.02572, 2023.

\bibitem[Dosovitskiy et~al.(2020)Dosovitskiy, Beyer, Kolesnikov, Weissenborn,
  Zhai, Unterthiner, Dehghani, Minderer, Heigold, Gelly,
  et~al.]{dosovitskiy2020image}
Dosovitskiy, A., Beyer, L., Kolesnikov, A., Weissenborn, D., Zhai, X.,
  Unterthiner, T., Dehghani, M., Minderer, M., Heigold, G., Gelly, S., et~al.
\newblock An image is worth 16x16 words: Transformers for image recognition at
  scale.
\newblock \emph{arXiv preprint arXiv:2010.11929}, 2020.

\bibitem[Everingham et~al.(2010)Everingham, Gool, Williams, Winn, and
  Zisserman]{EveringhamGWWZ10}
Everingham, M., Gool, L.~V., Williams, C. K.~I., Winn, J.~M., and Zisserman, A.
\newblock The pascal visual object classes {(VOC)} challenge.
\newblock \emph{Int. J. Comput. Vis.}, 88\penalty0 (2):\penalty0 303--338,
  2010.

\bibitem[Girdhar et~al.(2022)Girdhar, Singh, van~der Maaten, Joulin, and
  Misra]{GirdharSRMJM22}
Girdhar, R., Singh, M., van~der Maaten, L., Joulin, A., and Misra, I.
\newblock Omnivore: {A} single model for many visual modalities.
\newblock In \emph{{IEEE/CVF} Conference on Computer Vision and Pattern
  Recognition, {CVPR} 2022, New Orleans, LA, USA, June 18-24, 2022}, pp.\
  16081--16091. {IEEE}, 2022.

\bibitem[Goutte \& Gaussier(2005)Goutte and Gaussier]{goutte2005probabilistic}
Goutte, C. and Gaussier, E.
\newblock A probabilistic interpretation of precision, recall and f-score, with
  implication for evaluation.
\newblock In \emph{European conference on information retrieval}, pp.\
  345--359. Springer, 2005.

\bibitem[He et~al.(2016)He, Zhang, Ren, and Sun]{he2016deep}
He, K., Zhang, X., Ren, S., and Sun, J.
\newblock Deep residual learning for image recognition.
\newblock In \emph{2016 {IEEE} Conference on Computer Vision and Pattern
  Recognition, {CVPR} 2016, Las Vegas, NV, USA, June 27-30, 2016}, pp.\
  770--778. {IEEE} Computer Society, 2016.

\bibitem[He et~al.(2022)He, Chen, Xie, Li, Doll{\'a}r, and
  Girshick]{he2022masked}
He, K., Chen, X., Xie, S., Li, Y., Doll{\'a}r, P., and Girshick, R.
\newblock Masked autoencoders are scalable vision learners.
\newblock In \emph{Proceedings of the IEEE/CVF conference on computer vision
  and pattern recognition}, pp.\  16000--16009, 2022.

\bibitem[Hejna et~al.(2024)Hejna, Bhateja, Jiang, Pertsch, and
  Sadigh]{HejnaBJPS24}
Hejna, J., Bhateja, C.~A., Jiang, Y., Pertsch, K., and Sadigh, D.
\newblock Remix: Optimizing data mixtures for large scale imitation learning.
\newblock In Agrawal, P., Kroemer, O., and Burgard, W. (eds.), \emph{Conference
  on Robot Learning, 6-9 November 2024, Munich, Germany}, volume 270 of
  \emph{Proceedings of Machine Learning Research}, pp.\  145--164. {PMLR},
  2024.

\bibitem[Jocher et~al.(2020)Jocher, Stoken, Borovec, Changyu, Hogan, Diaconu,
  Poznanski, Yu, Rai, Ferriday, et~al.]{jocher2020ultralytics}
Jocher, G., Stoken, A., Borovec, J., Changyu, L., Hogan, A., Diaconu, L.,
  Poznanski, J., Yu, L., Rai, P., Ferriday, R., et~al.
\newblock ultralytics/yolov5: v3. 0.
\newblock \emph{Zenodo}, 2020.

\bibitem[Kirillov et~al.(2023)Kirillov, Mintun, Ravi, Mao, Rolland, Gustafson,
  Xiao, Whitehead, Berg, Lo, et~al.]{kirillov2023segment}
Kirillov, A., Mintun, E., Ravi, N., Mao, H., Rolland, C., Gustafson, L., Xiao,
  T., Whitehead, S., Berg, A.~C., Lo, W.-Y., et~al.
\newblock Segment anything.
\newblock In \emph{Proceedings of the IEEE/CVF international conference on
  computer vision}, pp.\  4015--4026, 2023.

\bibitem[Lin et~al.(2014)Lin, Maire, Belongie, Hays, Perona, Ramanan,
  Doll{\'a}r, and Zitnick]{lin2014microsoft}
Lin, T.-Y., Maire, M., Belongie, S., Hays, J., Perona, P., Ramanan, D.,
  Doll{\'a}r, P., and Zitnick, C.~L.
\newblock Microsoft coco: Common objects in context.
\newblock In \emph{Computer vision--ECCV 2014: 13th European conference,
  zurich, Switzerland, September 6-12, 2014, proceedings, part v 13}, pp.\
  740--755. Springer, 2014.

\bibitem[Ling et~al.(2020)Ling, Tollmar, and Gissl{\'e}n]{ling2020using}
Ling, C., Tollmar, K., and Gissl{\'e}n, L.
\newblock Using deep convolutional neural networks to detect rendered glitches
  in video games.
\newblock In \emph{Proceedings of the AAAI Conference on Artificial
  Intelligence and Interactive Digital Entertainment}, volume~16, pp.\  66--73,
  2020.

\bibitem[Loshchilov \& Hutter(2016)Loshchilov and Hutter]{loshchilov2016sgdr}
Loshchilov, I. and Hutter, F.
\newblock Sgdr: Stochastic gradient descent with warm restarts.
\newblock \emph{arXiv preprint arXiv:1608.03983}, 2016.

\bibitem[Loshchilov \& Hutter(2017)Loshchilov and
  Hutter]{loshchilov2017decoupled}
Loshchilov, I. and Hutter, F.
\newblock Decoupled weight decay regularization.
\newblock \emph{arXiv preprint arXiv:1711.05101}, 2017.

\bibitem[Oquab et~al.(2023)Oquab, Darcet, Moutakanni, Vo, Szafraniec, Khalidov,
  Fernandez, Haziza, Massa, El-Nouby, et~al.]{oquab2023dinov2}
Oquab, M., Darcet, T., Moutakanni, T., Vo, H., Szafraniec, M., Khalidov, V.,
  Fernandez, P., Haziza, D., Massa, F., El-Nouby, A., et~al.
\newblock Dinov2: Learning robust visual features without supervision.
\newblock \emph{arXiv preprint arXiv:2304.07193}, 2023.

\bibitem[Paduraru et~al.(2021)Paduraru, Paduraru, and
  Stefanescu]{cv_game_test21}
Paduraru, C., Paduraru, M., and Stefanescu, A.
\newblock Automated game testing using computer vision methods.
\newblock In \emph{2021 36th IEEE/ACM International Conference on Automated
  Software Engineering Workshops (ASEW)}, pp.\  65--72, 2021.

\bibitem[Qiu \& Yuille(2016)Qiu and Yuille]{unrealcv16}
Qiu, W. and Yuille, A.~L.
\newblock Unrealcv: Connecting computer vision to unreal engine.
\newblock In Hua, G. and J{\'{e}}gou, H. (eds.), \emph{Computer Vision - {ECCV}
  2016 Workshops - Amsterdam, The Netherlands, October 8-10 and 15-16, 2016,
  Proceedings, Part {III}}, volume 9915 of \emph{Lecture Notes in Computer
  Science}, pp.\  909--916, 2016.

\bibitem[Rahman(2023)]{rahman2023weak}
Rahman, F.
\newblock Weak supervision for label efficient visual bug detection.
\newblock \emph{arXiv preprint arXiv:2309.11077}, 2023.

\bibitem[Redmon et~al.(2016)Redmon, Divvala, Girshick, and Farhadi]{yolo16}
Redmon, J., Divvala, S.~K., Girshick, R.~B., and Farhadi, A.
\newblock You only look once: Unified, real-time object detection.
\newblock In \emph{2016 {IEEE} Conference on Computer Vision and Pattern
  Recognition, {CVPR} 2016, Las Vegas, NV, USA, June 27-30, 2016}, pp.\
  779--788. {IEEE} Computer Society, 2016.

\bibitem[Ren et~al.(2015)Ren, He, Girshick, and Sun]{girshick2015fast}
Ren, S., He, K., Girshick, R.~B., and Sun, J.
\newblock Faster {R-CNN:} towards real-time object detection with region
  proposal networks.
\newblock In \emph{Advances in Neural Information Processing Systems 28: Annual
  Conference on Neural Information Processing Systems 2015, December 7-12,
  2015, Montreal, Quebec, Canada}, pp.\  91--99, 2015.

\bibitem[Skwarczek(2021)]{game_ind_forbes}
Skwarczek, B.
\newblock How the gaming industry has leveled up during the pandemic. {F}orbes,
  2021.
\newblock [Online; Accessed: 2024-12-06].

\bibitem[Taesiri et~al.(2022)Taesiri, Macklon, Wang, Shen, and
  Bezemer]{taesirilarge}
Taesiri, M., Macklon, F., Wang, Y., Shen, H., and Bezemer, C.
\newblock Large language models are pretty good zero-shot video game bug
  detectors (2022), 2022.

\bibitem[Taesiri et~al.(2020)Taesiri, Habibi, and Fazli]{taesiri2020}
Taesiri, M.~R., Habibi, M., and Fazli, M.~A.
\newblock A video game testing method utilizing deep learning.
\newblock \emph{Iran Journal of Computer Science}, 17, 2020.

\bibitem[Taesiri et~al.(2024)Taesiri, Feng, Bezemer, and
  Nguyen]{taesiri2024glitchbench}
Taesiri, M.~R., Feng, T., Bezemer, C.-P., and Nguyen, A.
\newblock Glitchbench: Can large multimodal models detect video game glitches?
\newblock In \emph{Proceedings of the IEEE/CVF Conference on Computer Vision
  and Pattern Recognition}, pp.\  22444--22455, 2024.

\bibitem[Tamm et~al.(2022)Tamm, Shamon, Leon, Tollmar, and Gisslén]{tamm2022}
Tamm, M., Shamon, O., Leon, H.~A., Tollmar, K., and Gisslén, L.
\newblock Automatic testing and validation of level of detail reductions
  through supervised learning.
\newblock In \emph{2022 IEEE Conference on Games (CoG)}, pp.\  191--198, 2022.

\end{thebibliography}
\bibliographystyle{icml2023}



\end{document}